\documentclass[lettersize,journal]{IEEEtran}
\usepackage{amsmath,amsfonts}
\usepackage{algorithmic}
\usepackage{algorithm}
\usepackage{array}
\usepackage[caption=false,font=normalsize,labelfont=sf,textfont=sf]{subfig}
\usepackage{textcomp}
\usepackage{multirow}
\usepackage{stfloats}
\usepackage{amsmath}
\usepackage{amssymb}
\usepackage{amsthm}
\usepackage{txfonts}
\usepackage{url}
\usepackage{verbatim}
\usepackage{booktabs} 
\usepackage{graphicx}
\usepackage{cite}
\begin{document}

\title{VERHallu: Evaluating and Mitigating Event Relation Hallucination in Video Large Language Models}


\author{Zefan Zhang
\thanks{Zefan Zhang: zefan23@mails.jlu.edu.cn, College of Computer Science and Technology, Key Laboratory of Symbolic Computation and Knowledge Engineering, Ministry of Education, Jilin University.}, Kehua Zhu
\thanks{Kehua Zhu: zhukh25@mails.jlu.edu.cn, College of Software, Jilin University.}, Shijie Jiang
\thanks{Shijie Jiang: sjjiang25@mails.jlu.edu.cn, College of Computer Science and Technology, Key Laboratory of Symbolic Computation and Knowledge Engineering, Ministry of Education, Jilin University.}, 
Hongyuan Lu
\thanks{Hongyuan Lu: hongyuanlu@outlook.com, Facemind Group.}, 
Shengkai Sun \thanks{Shengkai Sun: ashenone1005@outlook.com, School of Computer Science and Information Engineering, Hefei University of Technology.}, 
Tian Bai* \thanks{Tian Bai*: Corresponding author. baitian@jlu.edu.cn, College of Computer Science and Technology, Key Laboratory of Symbolic Computation and Knowledge Engineering, Ministry of Education, Jilin University.}}

\markboth{}%
{VERHallu: Evaluating and Mitigating Event Relation Hallucination in Video Large Language Models}




\maketitle

\begin{abstract}

Video Large Language Models (VideoLLMs) exhibit various types of hallucinations. Existing research has primarily focused on hallucinations involving the presence of events, objects, and scenes in videos, while largely neglecting event relation hallucination. In this paper, we introduce a novel benchmark for evaluating the \textbf{V}ideo \textbf{E}vent \textbf{R}elation \textbf{Hallu}cination, named \textbf{VERHallu}. This benchmark focuses on causal, temporal, and subevent relations between events, encompassing three types of tasks: relation classification, question answering, and counterfactual question answering, for a comprehensive evaluation of event relation hallucination. Additionally, it features counterintuitive video scenarios that deviate from typical pretraining distributions, with each sample accompanied by human-annotated candidates covering both vision-language and pure language biases. Our analysis reveals that current state-of-the-art VideoLLMs struggle with dense-event relation reasoning, often relying on prior knowledge due to insufficient use of frame-level cues. Although these models demonstrate strong grounding capabilities for key events, they often overlook the surrounding subevents, leading to an incomplete and inaccurate understanding of event relations. To tackle this, we propose a Key-Frame Propagating (KFP) strategy, which reallocates frame-level attention within intermediate layers to enhance multi-event understanding. Experiments show it effectively mitigates the event relation hallucination without affecting inference speed. Our code and data is available at https://github.com/zefanZhang-cn/VERHallu.

\end{abstract}






\maketitle

\section{Introduction}

Large Language Models (LLMs) such as GPT \cite{gpt}, LLaMA \cite{llama,touvron2023llama}, and Qwen \cite{qwen} have demonstrated significant capabilities in natural language understanding and human-centered question solving. Recently, researchers have started incorporating visual encoders into LLMs with vision-language pretraining strategies \cite{li2023blip}, visual instruction tuning approaches \cite{liu2023visual,zhou2022learning}, and others to enhance the understanding of images and videos. These Multimodal Large Language Models (MLLMs) \cite{lin2024video,bai2025qwen2,chen2024internvl,zhu2023minigpt,wang2025align,yu2024prompting} with their powerful visual and textual understanding abilities have made remarkable progress in multimedia data mining from the social web \cite{luo2025imagescope,lin2024towards}, such as Sentiment Analysis and Fake News Detection.

\begin{figure*}
  \centering
\includegraphics[width=0.94\textwidth]{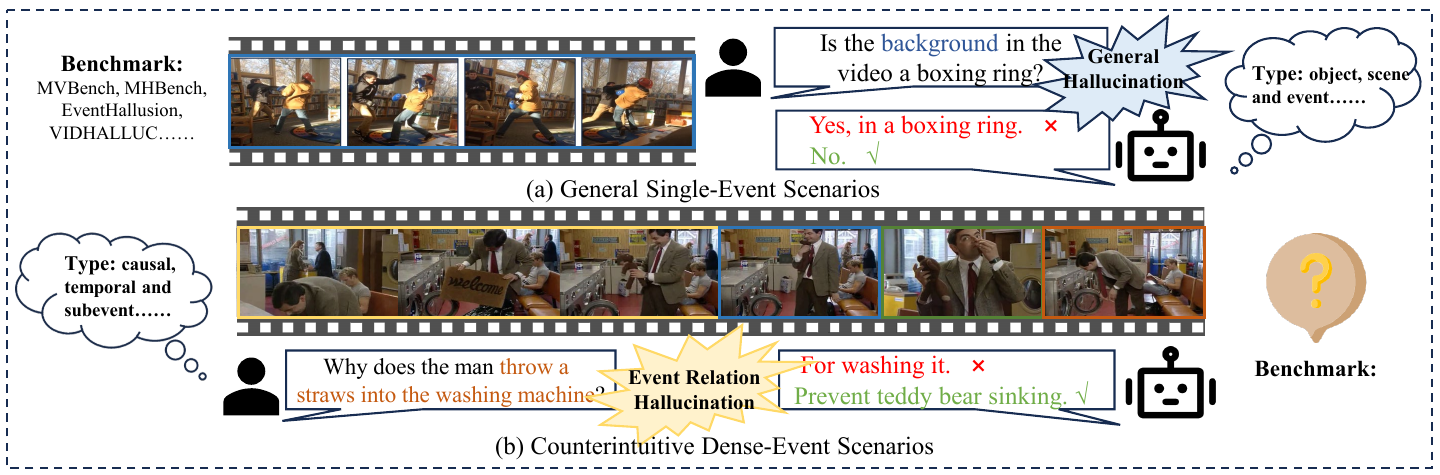} 
  \caption{Comparison of different hallucinations in VideoLLMs. Current video hallucination benchmarks focus on the existence of object, scene, and event in single-event scenarios, while ignoring the event relation hallucination in dense-event scenarios.} 
  \label{first}
\end{figure*}

Despite their impressive capabilities, LLMs and MLLMs suffer from hallucination issues, which refers to the model’s tendency to generate content that contradicts factual knowledge or visual evidence \cite{ji2023survey,liu2024survey,sahoo2024comprehensive,huang2024visual,guan2024hallusionbench,fang2024linguistic}. This issue is especially critical for MLLMs, as they are expected to accurately interpret visual inputs and generate responses that are faithful to the content of the image. Previous hallucination research has predominantly focused on verifying the existence of objects, events, or scenes in image and video contexts \cite{kong2025mhbench,gao2025exploring,li2023evaluating}, as shown in Figure \ref{first}. However, \textbf{they often overlook the more critical hallucination problem in dense video event settings, event relation hallucination}, as the comparison shown in Table \ref{tab:booktabs6}. Unlike existence-based hallucinations that can be mitigated by improving visual feature extractors to capture more comprehensive cues in single-event scenes \cite{qin2025question}, event relation hallucination requires deeper reasoning: \textbf{the model must not only recognize individual events, but also infer their relations by considering the scene context, object states, and the transitions among dense events}, as shown in Figure \ref{first}. This presents a greater challenge, pushing the boundaries of models’ deep and comprehensive understanding of video content. However, there are no comprehensive benchmarks specifically targeting this issue.

In this work, we propose \textbf{VERHallu}, a new benchmark for evaluating event relation hallucination in video large language models (VideoLLMs). Building upon prior work \cite{wang2022maven} in document-level event relation understanding, VERHallu focuses on a comprehensive set of event relation types, including \textbf{causal}, \textbf{temporal}, and \textbf{subevent}. To accurately reveal event relation hallucination while mitigating the influence of pretraining data distributions, we sample from the dense and counterintuitive event videos, which encompass a wide range of daily scenarios such as sports, dining, shopping, traveling, social interactions, and household activities, thereby providing representative contexts for evaluating event understanding. Then, we manually annotate and design challenging and disruptive candidate events with vision-language biases and language biases. Additionally, we design the benchmark with three task settings: (1) \textbf{Relation Classification (RC)}, where the model determines the relation (including directional causal, temporal, and subevent relations) between two events based on the video; (2) \textbf{Question Answering (QA)}, where the model answers the causal, temporal, or subevent based question from the biased candidates; and (3) \textbf{Counterfactual QA (CFQA)}, where the video is replaced with an unrelated one to test whether the model relies on textual priors rather than visual evidence. These tasks are complementary for a comprehensive evaluation. RC evaluates structured relation understanding, QA reflects complex practical event inference, and CFQA probes hallucination under conflicting visual context. The final dataset comprises 574 videos and 7676 manually verified samples.

We evaluate several recent state-of-the-art VideoLLMs on VERHallu and observe the following: (1) Existing VideoLLMs suffer from severe hallucination when handling dense video events. These models exhibit poor performance in classifying event relations, often performing near random chance. The QA accuracy of most models remains below 40\%. (2) VideoLLMs that achieve strong results on existing hallucination datasets show highly inconsistent and often poor performance in our complex dense-event scenarios, underscoring the unique value of our VERHallu dataset for revealing hallucinations. (3) Most VideoLLMs rely on static scene information (e.g., the first frame) and largely ignore the dynamic evolution of events, resulting in susceptibility to vision-language bias or language bias. We find that some VideoLLMs could accurately identify key events in the question but lack attention to nearby related events, which leads to hallucinations in event relations. To tackle this issue, we propose a Key-Frame Propagating (KFP) strategy, which reallocates frame-level attention around key frames within intermediate layers to enhance multi-event understanding. Experimental results show that KFP effectively mitigates hallucinations in video event relation understanding. The main contributions can be summarized as follows:
\begin{itemize}
    \item We present the first comprehensive analysis and benchmark, VERHallu, for evaluating the event relation hallucination in VideoLLMs.
    \item We propose a Key-Frame Propagating (KFP) strategy that effectively reduces the event relation hallucination while preserving inference efficiency, coherence, and accuracy.
    \item We conduct extensive empirical studies on advanced VideoLLMs with VERHallu and confirm that current models significantly suffer from the event relation hallucination, while our method shows clear improvements in mitigating this issue.
\end{itemize}




\begin{table*}[ht]
  \centering
  \caption{Comparison of VERHallu with recent hallucination benchmarks in video understanding. L: language bias. VL: vision\&language bias. ST: Spatial-Temporal. CS: Common Sense. Bias Type refers to the analysis of the sources of hallucinations during testing.}
  \label{tab:booktabs6}
  \begin{tabular}{l c c c c c}
    \toprule
    \textbf{Benchmarks} & \textbf{Data Source} & \textbf{Scenario Type} & \textbf{Dense Event} & \textbf{Bias Type} & \textbf{Hallucination Type} \\
    \midrule
    VIDHALLUC \cite{li2025vidhalluc} & Real & General & $\times$ &  N/A & Action, Temporal, Scene \\
    VideoHallucer \cite{wang2024videohallucer} & Real & General & $\times$  & N/A & Object, Temporal, Factual \\
    VideoHallu \cite{li2025videohallu} & Synthetic & General & $\times$  & N/A & Object, ST, CS, Physics \\
    HAVEN \cite{gao2025exploring} & Real & General& $\times$ & N/A & Object, Scene, Event \\
    UNSCENE \cite{bae2025mash} & Real & General & $\times$ & N/A & Action, Scene \\
    MHBench \cite{kong2025mhbench} & Synthetic & General & $\times$  & N/A & Motion \\
    EventHallusion \cite{Eventhallusion}& Real & General &  $\times$ & L, VL & Event \\
    \midrule
    VERHallu & Real & Counterintuitive & $\checkmark$ & L, VL & Event Relation \\
    \bottomrule
  \end{tabular}
\end{table*}

\section{Related Work}

\subsection{Hallucination Benchmark for VideoLLMs}

Compared to images, videos contain more dynamic and variable objects, events, and scenes, which makes hallucinations in VideoLLMs more diverse and complex. Eventhallusion \cite{Eventhallusion} focuses on diagnosing event existence hallucinations in VideoLLMs, specifically examining whether models falsely claim the presence or absence of certain events under language bias and vision-language bias settings. MHBench \cite{kong2025mhbench} introduces the notion of motion hallucination, revealing that models often misinterpret whether an action has occurred due to poor perception of dynamic changes across time. MASH-VLM \cite{bae2025mash} focuses on scene existence hallucinations, using GPT-generated labels to simulate visually grounded biases, while VideoHallu \cite{li2025videohallu} and VideoHallucer \cite{wang2024videohallucer} explore broader multimodal inconsistencies, including violations of physical laws and both intrinsic (e.g., temporal, semantic) and extrinsic (e.g., factual) hallucinations. VidHalluc \cite{li2025vidhalluc} further categorizes hallucinations into action, temporal sequence, and scene transitions, highlighting the importance of understanding video as a coherent temporal stream. Other works \cite{gao2025exploring} emphasize the influence of prior knowledge, contextual contradictions, and inherent model limitations across object, scene, and event levels. 


Despite these efforts, most existing benchmarks focus on the existence or sequence of actions, objects, or scenes, and ignore the more intricate problem of the event relation hallucination in video contexts. The whole comparison of different hallucination datasets is shown in Table \ref{tab:booktabs6}.


\subsection{Mitigating Hallucination in VideoLLMs}

For addressing the motion hallucination, Kong et al. \cite{kong2025mhbench} leverage bidirectional video pairs comparing forward and reversed action sequences to disambiguate motion perception through visual contrastive decoding \cite{vcd}. Wang et al. \cite{wang2024videohallucer} propose the integration of synthetic video data with Group Relative Policy Optimization (GRPO) to enhance the model’s capacity for abnormality detection and reasoning. Furthermore, hallucinations are particularly pronounced in long video understanding tasks, where models tend to overlook relevant content due to limited memory and attention spans. To mitigate such issues, methods \cite{sun2024hallucination} incorporating CLIP-based frame sampling, question-guided frame selection, and controlled generation strategies have been proposed to better align model outputs with the actual content of extended video sequences. PaMi-VDPO \cite{ding2025pami} advances this line of work by incorporating prompt-aware multi-instance preference learning, enabling the model to adaptively filter noisy samples and align outputs with diverse prompt conditions. VISTA-LLAMA \cite{ma2024vista} incorporates an equal-distance attention mechanism between visual and textual tokens to preserve visual context throughout long-form generation, whereas DINO-HEAL \cite{li2025vidhalluc} integrates spatial saliency from DINOv2 \cite{oquab2023dinov2} to reweight visual features during inference to suppress action and temporal hallucinations.


\begin{figure*}
  \centering
\includegraphics[width=0.9\textwidth]{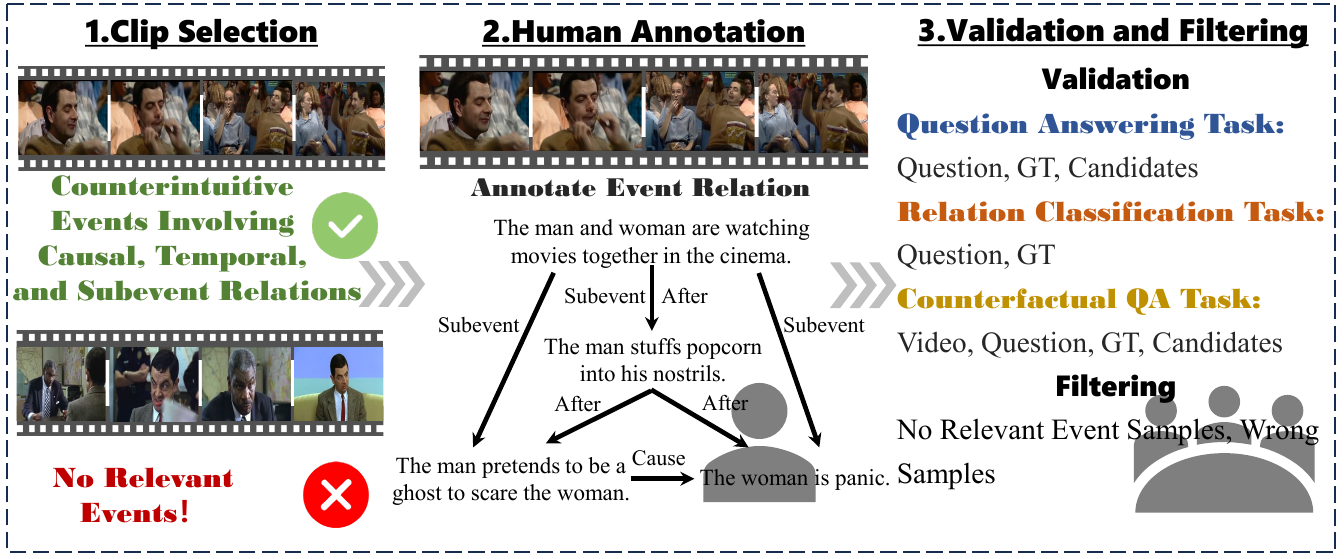} 
  \caption{Illustration of VERHallu construction process. V\&L bias: vision\&language bias. L bias: language bias. GT: ground truth.} 
  \label{label}
\end{figure*}

\section{Benchmark: VERHallu}

In this section, we first introduce the construction process of VERHallu and then demonstrate how to define the three types of tasks for evaluating the event relation hallucination in VideoLLMs.

\begin{figure}
  \centering
\includegraphics[width=0.45\textwidth]{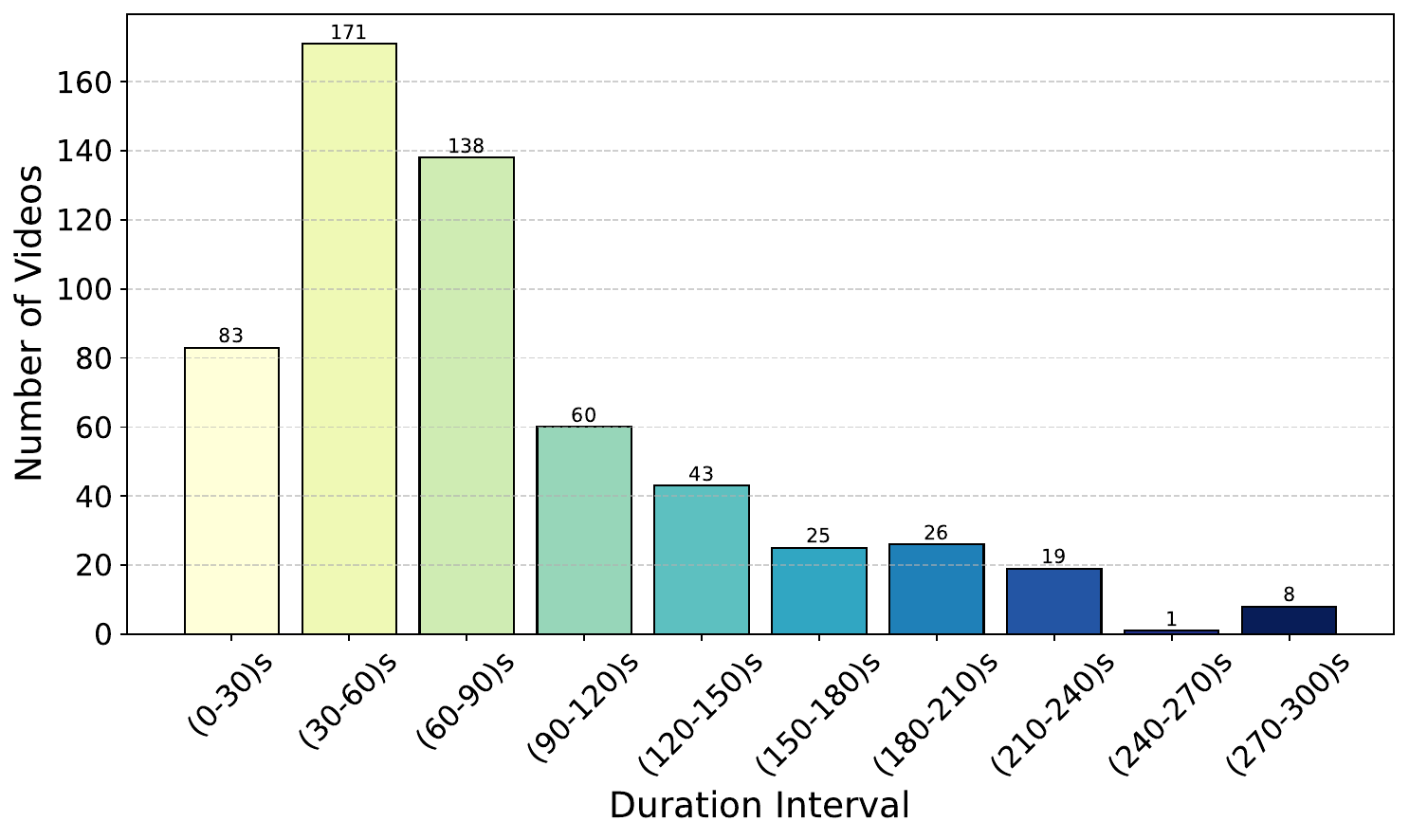} 
  \caption{The distribution of video durations in VERHallu.} 
  \label{duration}
\end{figure}

\subsection{Data Collection}
Previous benchmarks for evaluating video hallucinations have typically relied on either existing datasets \cite{kong2025mhbench,li2025vidhalluc} or synthetically generated videos \cite{bae2025mash,li2025videohallu} designed to exhibit specific types of hallucinations. However, such data often suffer from limitations: they either closely resemble the distribution of visual instruction tuning datasets, leading to overfitting, or lack realism and accuracy, making it difficult to faithfully reflect the hallucination issues faced by VideoLLMs. In particular, when it comes to hallucinations involving event relations, traditional datasets or construction methods fall short in evaluating the complex dependencies between events.

To address these limitations, we carefully curated a dataset based on ``Mr. Bean'' videos, which feature event sequences that appear counterintuitive at first glance, yet are grounded in real-world logic. These videos naturally embed causal, temporal, and subevent relations that can be accurately interpreted through contextual analysis. Moreover, their counterfactual nature effectively reduces the influence of language or vision-language biases, enabling a more precise evaluation of VideoLLMs’ event relation hallucination.

We selected ``Mr. Bean Episodes 1 to 14'', ``Mr. Bean's Holiday'', and ``Bean: The Ultimate Disaster Movie'' as our video data sources, with a total duration of approximately 25 hours.

\subsection{Data Annotation}

Our dataset annotation process consists of several key steps, primarily focusing on extracting video segments that feature counterintuitive yet contain causal, temporal, or subevent relations. Each selected clip lasts approximately one to two minutes. For each segment, questions are formulated based on the counterintuitive events.

\begin{figure*}
  \centering
\includegraphics[width=0.9\textwidth]{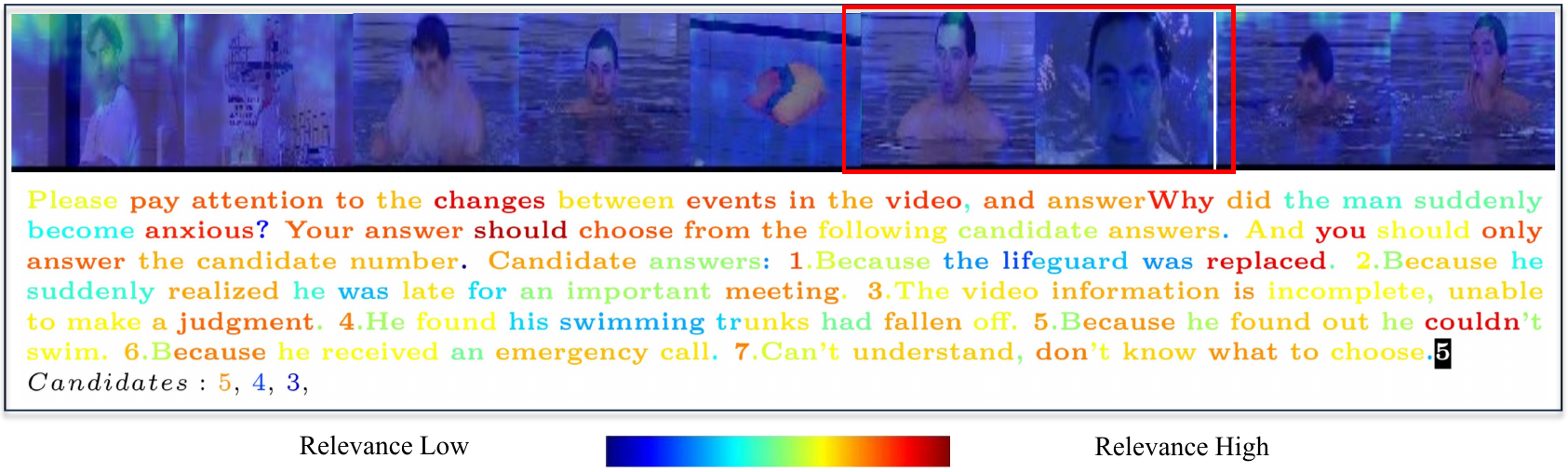} 
  \caption{Visualization results of the QwenVL-2.5-7B with TAM.} 
  \label{visuallll}
\end{figure*}

\textbf{Question Answering (QA) Task: }For temporal question answering, annotators create questions phrased as: ``What happens after [Event]?''. For causal question answering, the format is: ``Why does the person perform [Event]?''. For subevent question answering, the format is: ``During [Main\_Event], which of the following event occurred?''. Annotators then provide the correct answer and construct two types of biased candidates: two candidates based on vision-language priors (using both video and question context), and two candidates based on language priors (using only the question).

\textbf{Relation Classification (RC) Task: }This task requires annotators to label the relation between pairs of events, drawing from three categories of event relations: causal, temporal, and subevent. Each relation type is annotated with direction, resulting in three candidate labels for the relation classification task, such as \textit{none}, \textit{cause}, and \textit{effect}.

\textbf{Counterfactual Question Answering (CFQA) Task: }We adapt the original QA questions to video segments in which the queried event does not occur. During the validation phase, we ensure that the target event is indeed absent. Candidate answers in this setting include ``Video information is incomplete, unable to judge.'' and ``I can't understand, don't know what to choose.''. This task is designed to assess the model’s ability to recognize uncertainty and avoid hallucinated responses when presented with insufficient or misleading visual information.

\subsection{Data Validation and Filtering}

For each sample, a two-stage annotation process was employed. After the initial annotation, a second annotator was introduced to verify and review the sample. In cases where disagreement arose between the two annotators, a third annotator was brought in to make the final judgment. If consensus could not be reached, the sample was discarded. Among the 967 QA samples, a total of 43 errors were identified and corrected. For the 967 counterfactual QA samples, all videos meet the requirements. In the case of the 5742 relation classification samples, 154 errors were identified, and 10 samples were discarded due to unreasonable content. The Cohen’s k QA: 86.2\%, CFQA:  99.9\%, RC: 85.8\%.

\subsection{Dataset Statistics}

Finally, we collect 7676 samples from 574 video clips, including 967 QA samples, 967 counterfactual QA samples, and 5742 relation classification samples. In QA samples, there are 212 temporal type questions, 497 causal type questions, and 258 subevent type questions. The distribution of video durations in VERHallu is shown in Figure \ref{duration}.

In the relation classification task, we construct three types of relation questions: temporal, causal, and subevent. The distribution of samples across relation types and their respective labels is as follows:

\begin{itemize}
    \item \textbf{Temporal relation:} A total of 2,683 samples, including:
    \begin{itemize}
        \item \textit{Before:} 665 \textit{After:} 669 \textit{None:} 1,349
    \end{itemize}

    \item \textbf{Causal relation:} A total of 1,511 samples, including:
    \begin{itemize}
        \item \textit{Cause:} 135 \textit{Effect:} 138 \textit{None:} 1,238
    \end{itemize}

    \item \textbf{Subevent relation:} A total of 1,548 samples, including:
    \begin{itemize}
        \item \textit{Main\_Event:} 258 \textit{Sub\_Event:} 258 \textit{None:} 1,032
    \end{itemize}
\end{itemize}

These statistics highlight the diversity and depth of VERHallu, underscoring its value as a comprehensive benchmark for evaluating video event relation hallucination in VideoLLMs.

\section{Method}
In this section, we analyze the issue of event relation hallucination and propose a training-free Key-Frame Propagating strategy.
\subsection{Motivation}

Building upon the Token Activation Map (TAM) method \cite{li2025token}, we visualize the final layer token activations of the QwenVL-2.5-7B \cite{qwen} to interpret its visual attention during generation, as shown in Figure \ref{visuallll}. 

The results reveal that QwenVL-2.5-7B tends to concentrate on the early frames of the video. Although QwenVL-2.5-7B demonstrates relatively strong visual grounding capabilities and partially attends to frames that depict the man's anxiety, it fails to attend adequately to frames near the key moments. This insufficient focus on critical visual cues (the red box) leads to incorrect inference of event relations and ultimately results in erroneous responses. Additionally, this phenomenon is also substantiated in Figure \ref{bias_analysis}, where the model demonstrates a tendency to rely on incorrect or incomplete visual information, thereby exhibiting more serious vision\&language bias. This issue can be attributed to the data distribution during pretraining. Specifically, the pretraining process emphasizes the vision\&language alignment and event or entity grounding while providing limited training for complex event relation reasoning. As a result, the model exhibits these deficiencies, which become especially obvious in videos with dense and complex event structures.

\subsection{Key-Frame Propagating Strategy}

Based on the aforementioned observations, we propose a \textbf{Key-Frame Propagation (KFP)} strategy, which enhances the saliency of surrounding frames by leveraging the temporal influence of the most informative frames through a Gaussian distribution mechanism.

Specifically, similar to prior work \cite{yin2025clearsight}, we select the hidden states from layers 8 to 15 for feature propagation. For each layer's hidden states $H$ before sending to the MLP layer, we extract the visual token representations $\mathbf{V} \in \mathbb{R}^{T\times N \times D}$, where $T$ denotes the number of frames, $N$ is the token number of each frame, and $D$ is the feature dimension. We then rank the attention score to select the top-$k$ key frames $\{t_1^*, t_2^*, \dots, t_k^*\}$.

For each selected key frame $t_i^*$, a temporal Gaussian weighting is applied over a window of $m$ frames to propagate its attention score to adjacent frames. The propagated attention for a neighboring frame $t$ is defined as:
\begin{equation}
\alpha_t = \exp\left(-\frac{(t - t_i^*)^2}{2\sigma^2}\right),
\end{equation}
where $\sigma$ controls the spread of the attention. Then, we integrate all top-$k$ results and get $\alpha$ to enhance the visual feature $\mathbf{V}$ in the hidden state feature $H$: 
\begin{equation}
\mathbf{V}_E= Softmax(\alpha+1) * \mathbf{V}.
\end{equation}

Subsequently, the enhanced visual features are concatenated with the original textual features to obtain the visual-enhanced hidden state feature $H_E$. Finally, the Multi-Layer Perceptron (MLP) layer is employed to fuse the enhanced hidden representations:
\begin{equation}
\mathbf{H}=(1-\beta)* H_E+\beta *H,
\end{equation}
where $\beta$ is a hyper-parameter.



This strategy is entirely training-free and introduces negligible computational overhead during inference. It allows the model to better capture event-centric temporal dependencies by focusing more holistically on the contextual relationships around key visual events.

\begin{table*}[ht]
\centering
 \caption{Accuracy comparison of different models on MVBench and VERHallu (\%). CFQA: Counterfactual Question Answering task, QA: Question Answering task, RC: Relation Classification task, C: Causal, T: Temporal, S: Subevent.  The bolded result represents the best outcome. The best results are highlighted in bold.}
\begin{tabular}{l cccccccccc}
\toprule
\multirow{2}{*}{\textbf{Models}} & \multirow{2}{*}{\textbf{Frames}} & \multirow{2}{*}{\textbf{MVBench}} & \multirow{2}{*}{\textbf{CFQA}} & 
\multicolumn{3}{c}{\textbf{QA}} & 
\multicolumn{3}{c}{\textbf{RC}} & 
\multirow{2}{*}{\textbf{SRH}} \\
\cmidrule(lr){5-10}
& & & & \textbf{C} & \textbf{T} & \textbf{S} & \textbf{C} & \textbf{T} & \textbf{S} & \\
\midrule
Random &  - & 27.3 & 14.3 & 14.3 & 14.3 & 14.3 & 33.3 & 33.3 & 33.3 & - \\
\midrule
ShareGPT4Video-8B \cite{chen2024sharegpt4video} &fps=0.5& -&4.4 &21.7 &24.1 &28.7 &8.9 &25.5 &14.9 & 7.6 \\
MiniCPM-V-2\_6-7B \cite{yao2024minicpm} & 12&- &36.4 &32.9 &33.9 & 33.3&14.1 &27.6 &16.8 &24.6 \\
Video-LLaVA-7B \cite{lin2024video}&8 &- & 3.3&12.1 &13.7 &11.2 &12.0 &24.9 &17.1 & 6.5 \\
LLaVA-NeXT-7B \cite{li2024llava}  & 32&- &2.7&15.0 &15.0 &12.0 &9.4 &24.7 &16.7 &5.7 \\
LLaVA-NeXT-7B-DPO \cite{li2024llava}&32 &40.0 &2.9 &17.9 &17.0 &16.7 &8.6 &24.9 &13.9 & 4.9 \\
LLaVA-Onevision-7B \cite{Llava-onevision}   &64 &56.7 & 13.3&30.8 &34.4 &39.5 &10.1 &24.9 &16.3 & 19.9 \\
mPLUG-Owl3-7B(0728) \cite{ye2024mplug} & 16& 54.5&15.8 &39.2&39.2 &33.7 &11.4&25.5 &16.2 &20.3 \\
mPLUG-Owl3-7B(1101) \cite{ye2024mplug}  & 16& 59.5&20.6 &37.6 &36.8 &36.4 & 11.9&22.3 &15.4 & 20.5\\
VideoLLaMA3-7B \cite{zhang2025videollama} &fps=1 &67.7 &31.9 & 15.9&20.8& 19.4&\textbf{47.3} &\textbf{40.5} &\textbf{44.1} & \textbf{38.7} \\
VideoChat-R1-7B \cite{li2025videochat} &fps=1  & 67.9&9.4 & 40.8&38.7 &38.4 &11.7 &25.1 &18.7 & 22.6 \\
QwenVL-2-7B \cite{wang2024qwen2} &fps=1 &67.0&28.1&\textbf{41.9} &36.8 &36.4 &11.4 &23.9 &17.6 & 22.9 \\
QwenVL-2.5-7B \cite{bai2025qwen2} &fps=1 & 69.6&\textbf{50.6} &40.8 &\textbf{50.5} &\textbf{40.7} &14.9 &24.9 &20.6 & 27.8 \\
InternVL-2.5-8B \cite{chen2024internvl} & 8&72.0 &29.3 &28.4 &27.4&32.9 &22.1 &30.9 &24.2 & 26.4\\
InternVL-3-8B \cite{chen2024internvl} &8 &\textbf{75.4} &15.6 &35.4 & 42.9& 36.4&26.8 &28.0 &30.2 & 27.6 \\
\bottomrule
\end{tabular}

\label{tab:models_comparison1}
\end{table*}


\section{Experiments}

We evaluate the event relation hallucination performance on VERHallu using 14 recent state-of-the-art open-source VideoLLMs. We choose the 7B or 8B model scale while preserving each baseline's original settings to ensure fairness. All models are evaluated on 1 NVIDIA A40 with 48G memory. For the QA and CFQA tasks, we primarily use accuracy as the evaluation metric. In the relation classification task, we employ precision, recall, F1-score, and accuracy to assess the models comprehensively. We further propose the Structure Relation Hallucination (SRH) metric, which calculates the average accuracy per video by first computing the proportion of correctly answered questions within each video (including CFQA, QA, and RC), and then averaging these accuracies across all videos.

\section{Implementation Details}
\label{detail}

\textbf{MVBench} \cite{li2024mvbench} is a comprehensive multi-modal video understanding benchmark designed to evaluate the temporal reasoning abilities of Multi-modal Large Language Models (MLLMs). Unlike existing benchmarks that focus on static image tasks, MVBench covers 20 dynamic video tasks requiring temporal comprehension beyond a single frame. Using a novel static-to-dynamic task conversion method, MVBench systematically generates video-based multiple-choice QA tasks from public annotations. This ensures fair and scalable evaluation with minimal manual effort. In our implementation of the VCD method, we followed the approach proposed in these works \cite{kong2025mhbench,vcd} and used the same hyperparameter settings to ensure consistency.

More details of the Video Large Language Models in our work:

\begin{itemize}
    \item InternVL-3-8B \cite{chen2024internvl} stands out in video understanding by leveraging unified multimodal pre-training and V2PE, allowing it to model long-range temporal dependencies and complex visual-language interactions more effectively than traditional MLLMs. 
    \item QwenVL-2.5-7B \cite{bai2025qwen2} excels in video understanding with long-duration processing, precise temporal localization, and dynamic resolution handling, enabling robust performance on complex real-world video tasks. 
    \item Video-LLaVA-7B \cite{lin2024video} unifies visual representation into the language feature space to advance the foundational LLM towards a unified LVLM. 
    \item VideoChat-R1-7B \cite{li2025videochat} is a video-based multimodal large language model fine-tuned with reinforcement learning to enhance spatio-temporal reasoning while maintaining strong general and conversational capabilities.
    \item VideoLLaMA3-7B \cite{zhang2025videollama} is a vision-centric multimodal foundation model designed with a high-quality image-text training paradigm and adaptive framework to achieve strong performance in both image and video understanding tasks.
    \item LLaVA-NeXT-Video-7B \cite{li2024llava} extends its image-trained capabilities to video understanding through techniques like AnyRes and length generalization, achieving strong zero-shot performance and scalable inference, with further gains via supervised and preference-aligned fine-tuning. LLaVA-NeXT-Video-7B-DPO \cite{li2024llava} is trained with Direct Preference Optimization.
    \item LLaVA-Onevision-7B \cite{Llava-onevision} with design in modeling and data representations allows task transfer across different scenarios, suggesting a simple approach to yield new emgerging capabilities.
    \item ShareGPT4Video-8B \cite{chen2024sharegpt4video} is a simple yet superb LVLM that reached SOTA performance on three advancing and comprehensive video benchmarks.
    \item mPLUG-Owl3-7B \cite{ye2024mplug} is a versatile multimodal large language model designed for robust long image-sequence and video understanding through hyper attention blocks, achieving state-of-the-art performance across diverse benchmarks and excelling in distractor-resistant visual reasoning.
    \item MiniCPM-V-2\_6-7B \cite{yao2024minicpm} is an efficient and compact multimodal large language model series optimized for end-side deployment, achieving GPT-4V-level performance across benchmarks while supporting high-resolution perception, multilingual understanding, and mobile inference with low computational cost.
    \item ChatGPT-4o \cite{bubeck2023sparks} and Gemini-3-Pro \cite{team2023gemini} are closed-sourced models with the strong performance. Due to cost, we randomly sampled 200 examples per question type for ChatGPT-4o and Gemini-3-Pro. 
\end{itemize}

\subsection{Prompts for Different Tasks}
\label{prompt}

We design specific prompts for different relation classification and question answering tasks. Each prompt instructs the model to make a choice from a predefined set of candidate answers by returning only the corresponding number. The templates for each task are as follows:

\begin{itemize}

    \item \textbf{Relation Classification (Causal)} \\
    \begin{enumerate}
        \item \texttt{According to the video, [question] Your answer should choose from the following candidate answers. You should only answer the candidate number.}
        \item \texttt{Candidate answers: (1) None (2) Cause (3) Effect.}
        \item \texttt{Cause: Event A causes Event B. Effect: Event B causes Event A.}
    \end{enumerate}

    \item \textbf{Relation Classification (Temporal)} \\
    \begin{enumerate}
        \item \texttt{According to the video, [question] Your answer should choose from the following candidate answers. You should only answer the candidate number.}
        \item \texttt{Candidate answers: (1) None (2) Before (3) After.}
        \item \texttt{Before: Event B occurs before Event A. After: Event A occurs before Event B.}
    \end{enumerate}

    \item \textbf{Relation Classification (Subevent)} \\
    \begin{enumerate}
        \item \texttt{According to the video, [question] Your answer should choose from the following candidate answers. You should only answer the candidate number.}
        \item \texttt{Candidate answers: (1) None (2) Main\_Event (3) Sub\_Event.}
        \item \texttt{Main\_Event: Event A contains Event B. Sub\_Event: Event B contains Event A.}
    \end{enumerate}

    \item \textbf{Question Answering} \\
    \begin{enumerate}
        \item \texttt{According to the video, [question] Your answer should choose from the following candidate answers. You should only answer the candidate number.}
        \item \texttt{Candidate answers: (1) [candidate\_1] (2) [candidate\_2] (3) [candidate\_3] (4) [candidate\_4] (5) [candidate\_5] (6) [candidate\_6] (7) [candidate\_7]}
    \end{enumerate}

\end{itemize}

\begin{table*}
    \centering
    \caption{Additional comparison results on VERHallu Relation Classification task (\%). P: precision, R: Recall. The best results are highlighted in bold.}
    \begin{tabular}{lccccccccc}
        \toprule
        \multirow{2}{*}{Models} & \multicolumn{3}{c}{\textbf{RC-Causal}} & \multicolumn{3}{c}{\textbf{RC-Temporal}} & \multicolumn{3}{c}{\textbf{RC-Subevent}} \\ 
        \cmidrule(lr){2-10}
                                & P & R & F1 & P & R & F1 & P & R & F1 \\ \midrule
ShareGPT4Video-8B \cite{chen2024sharegpt4video} &8.8& 48.7&14.9 &25.5 &\textbf{51.2} &34.0 &14.6 &43.6 &21.9 \\
MiniCPM-V-2\_6-7B \cite{yao2024minicpm} & 9.3&48.4 &15.6 &25.7 &50.3 & \textbf{34.1}&\textbf{16.7} &25.0 &16.8\\
Video-LLaVA-7B \cite{lin2024video}&8.5&44.7 &14.3 &24.9 &50.1 &33.3 &16.5 &\textbf{49.0} &\textbf{24.7}\\
LLaVA-NeXT-7B \cite{li2024llava}  & 8.1&43.9 &13.7 &24.7 &49.7 &33.0 &15.7 &46.3 &23.4 \\
LLaVA-NeXT-7B-DPO \cite{li2024llava}&8.6&47.6 &14.6 &24.9 &33.2 &24.9 &13.9 &41.7 &20.8\\
LLaVA-Onevision-7B \cite{Llava-onevision}   &8.7 &46.9& 14.6&24.9 &50.0 &33.2 &16.3 &48.8 &24.4 \\
mPLUG-Owl3-7B(0728) \cite{ye2024mplug} & 8.9& 47.6&15.1 &25.0 &49.9 &33.3 &15.9 &47.7 &23.9 \\
mPLUG-Owl3-7B(1101) \cite{ye2024mplug}  & 10.0& \textbf{53.8}&\textbf{16.8} &21.9 &43.7 &29.2 & 13.6&39.7 &20.2 \\
VideoLLaMA3-7B \cite{zhang2025videollama} &\textbf{10.6} &28.6 &15.5 & \textbf{26.0}&20.9& 23.2&15.0 &19.8 &17.1 \\
VideoChat-R1-7B \cite{li2025videochat} &8.5 &45.4&14.4 &24.7 &49.5 &32.8 &14.8 &41.9 &21.8\\
QwenVL-2-7B \cite{wang2024qwen2} &9.2 & 49.8&15.6 &23.6 &47.3 &31.5 &15.8 &45.7 &17.6 \\
QwenVL-2.5-7B \cite{bai2025qwen2} &9.7 & 49.8&16.2 &24.1 &47.9 &32.0 &14.6 &40.3 &21.5 \\
InternVL-2.5-8B \cite{chen2024internvl} & 7.8&34.8 &12.8 &25.7 &43.6 &32.3 &13.6 &32.4 &19.1 \\
InternVL-3-8B \cite{chen2024internvl} &9.1 &38.1 &14.6 &24.5 & 44.0& 31.5&16.4 &37.0 &22.7 \\
\bottomrule
    \end{tabular}
    
    \label{tab:models_comparison2}
\end{table*}

\begin{table*}[ht]
\centering
\caption{Accuracy comparison of large-scale Video LLMs on VERHallu (\%).}
\begin{tabular}{l ccccccc}
\toprule
\textbf{Models} & \textbf{CFQA} & \textbf{QA-Causal} & \textbf{QA-Temp} & \textbf{QA-Subevent} & \textbf{RC-Causal} & \textbf{RC-Temp} & \textbf{RC-Subevent} \\
\midrule
Human & 96.8 & 94.9 & 92.4 & 93.8 & 86.9 & 85.2 & 87.6 \\
ChatGPT-4o \cite{bubeck2023sparks}& 68.5 & 46.5 & 36.0 & 42.5 & 24.5 & 32.0 & 33.5 \\
Gemini-3-Pro \cite{team2023gemini} & 45.0 & 65.7 & 55.0 & 40.0 & 30.0 & 35.0 & 40.0 \\
Qwen-VL-2.5-32B \cite{bai2025qwen2}& 44.9 & 39.6 & 34.4 & 31.4 & 16.3 & 18.0 & 29.6 \\
Qwen-VL-2.5-72B \cite{bai2025qwen2}& 67.6 & 28.9 & 28.7 & 34.8 & 31.0 & 34.1 & 33.4 \\
\bottomrule
\end{tabular}
\label{tab:llllm}
\end{table*}

\subsection{Results on Main Baselines}

\textbf{Overall Performance on VERHallu: }In the Table \ref{tab:models_comparison1}, VERHallu presents a more significant challenge for current VideoLLMs. On both standard Question Answering (QA) and CounterFactual QA (CFQA) tasks, few models achieve accuracy above 50\%, with the exceptions of QwenVL-2.5-7B in CFQA and QA-T. Furthermore, certain models, such as ShareGPT4Video-8B and LLaVA-NeXT-7B, exhibit extremely low accuracy on the CFQA task. In the relation classification task, all models except VideoLLaMA3-7B perform worse than a random guess. Moreover, as shown in Table \ref{tab:llllm}, larger-scale models are introduced for evaluation. The results indicate that even with increased model size, existing methods still suffer from severe event relational hallucination issues.


\textbf{Comparison Between MVBench and VERHallu: }MVBench \cite{li2024mvbench} primarily evaluates fundamental visual understanding in video models. However, \textbf{stronger performance on MVBench does not translate to better results on VerHallu.} For example, InternVL-3-8B and QwenVL-2.5-7B rank among the top performers on MVBench but exhibit poor performance on VERHallu-CFQA or VERHallu-RC. The findings underscore a pronounced distribution gap between the VERHallu dataset and traditional benchmarks, suggesting that existing datasets fall short in evaluating deeper video reasoning abilities.

\textbf{Model Comparisons Across Sub-tasks: }Across sub-tasks, notable differences in model performance emerge. These three sub-tasks together provide a comprehensive evaluation from different dimensions of event relation understanding in VideoLLMs. QA assesses fundamental understanding but still allows models to succeed through textual priors. CFQA better reflects whether the model attends to video specific information, though it focuses less on relation logic. Relation classification imposes stricter requirements by specifying relation directions, offering a deeper test of event relation understanding.

QwenVL-2.5-7B demonstrates strong performance in both the CFQA and QA tasks, largely attributed to its robust capabilities in event and entity localization. As illustrated in the Figure \ref{visuallll}, the model can accurately identify the events mentioned in the input text and determine their presence in videos, thereby alleviating partial hallucination issues. However, when it comes to reasoning about relations between events, the model struggles due to its limited attention to relational information, resulting in a higher incidence of relation-level hallucinations, for instance, its relatively poor performance in RC.

VideoLLaMA3-7B performs notably well in the RC task, being the only model to surpass the random-guess baseline across all three relation types consistently. Nevertheless, as shown in Table \ref{tab:models_comparison2}, it achieves relatively lower F1 scores, suggesting that while it correctly identifies more ``None'' relation samples, its reasoning over event relations remains superficial. Meanwhile, the model also demonstrates superior capability in understanding holistic video event relations (SRH 38.7\%).

InternVL-3-8B achieves a more balanced performance, with both high overall accuracy and strong F1 scores in RC. This highlights its superior analytical capabilities. However, considering its worse performance in the CFQA task, it may exhibit certain language biases that affect its general reasoning ability.

Results taken together of these sub-tasks reveal that current VideoLLMs fail to meet expectations in all three dimensions of event relation understanding, and are thus unable to provide users with accurate and comprehensive video comprehension.

\begin{figure}
  \centering
\includegraphics[width=0.45\textwidth]{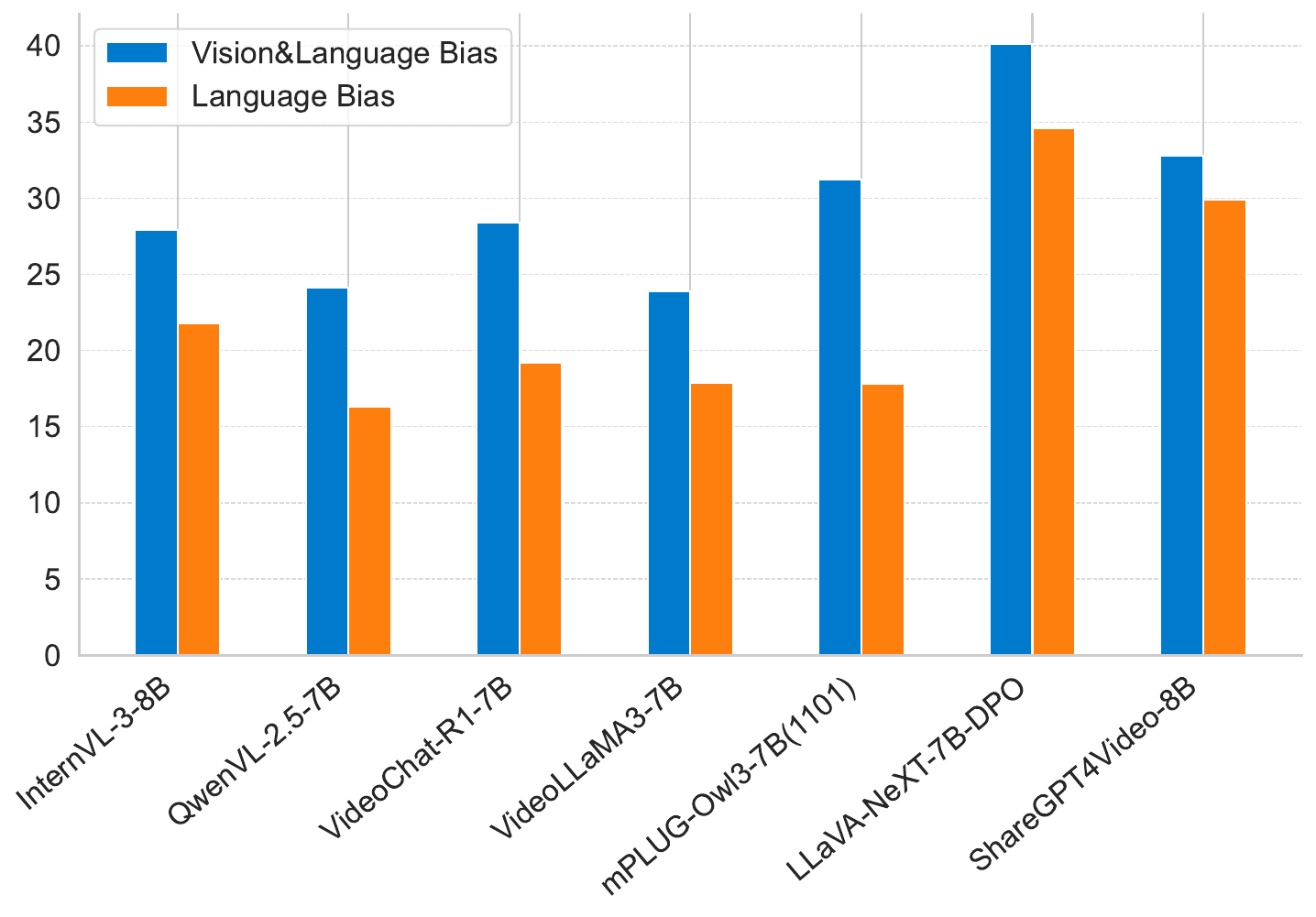} 
  \caption{Vision\&Language Bias and Language Bias analysis (\%).} 
  \label{bias_analysis}
\end{figure}

\begin{table*}[ht]
\centering
\caption{Experimental results of KFP in VERHallu.}
\begin{tabular}{l ccccccccc}
\toprule
\multirow{2}{*}{\textbf{Models}}  & \multirow{2}{*}{\textbf{CFQA}} & 
\multicolumn{3}{c}{\textbf{QA}} & 
\multicolumn{3}{c}{\textbf{RC}} & 
\multirow{2}{*}{\textbf{SRH}} & 
\multirow{2}{*}{\textbf{Inference Speed (FPS)}}\\
\cmidrule(lr){3-8}
 & & \textbf{C} & \textbf{T} & \textbf{S} & \textbf{C} & \textbf{T} & \textbf{S} & & \\
\midrule
LLaVA-NeXT-7B &2.7 &15.0 &15.0 &12.0 &9.4 &24.7 &16.7 & 6.5 & 0.070\\
LLaVA-NeXT-7B + KFP &9.1 &15.3 &19.3 & 15.1 &9.4 &27.5 &55.8 & 17.9 & 0.067\\ 
\midrule
QwenVL-2.5-7B &50.6 &40.8 &50.5 &40.7 &14.9 &24.9 &20.6 & 27.8 & 0.142 \\
QwenVL-2.5-7B + VCD &48.5 &41.9 &49.5 &38.3 &13.5 &24.4 &19.9 & 18.7 & 0.138 \\
QwenVL-2.5-7B + KFP &44.6 &41.9 &49.1 &45.3 &15.2 &24.6 &21.8 & 43.5 & 0.124\\ 
\bottomrule
\end{tabular}

\label{tab:models_kfp}
\end{table*}

\textbf{Event Relation Types: }In Table \ref{tab:models_comparison1}, we present the evaluation results of VideoLLMs across three event relation types within the Question Answering and Relation Classification tasks. While causal reasoning is intuitively regarded as more complex than temporal or subevent reasoning, our dataset intentionally includes counterintuitive event pairs to elevate the task difficulty. Consequently, models continue to exhibit worse performance even on temporal and subevent relation questions. This further underscores their reliance on superficial cues and highlights the prevalence of event relation hallucination. Moreover, most models demonstrate notably weaker performance on RC-Causal, with RC-Subevent and RC-Temporal relations performing slightly better.


\textbf{Number of Frames:} Table \ref{tab:models_comparison1} compares the frame sampling strategies used by different models. Models like InternVL-3-8B, which process fewer frames, perform significantly worse on the CFQA task. In contrast, models such as VideoLLaMA2-7B and VideoChat-R1-7B sample more frames but do not show corresponding performance gains. QwenVL-2.5-8B, which combines an effective sampling strategy with strong event localization, achieves state-of-the-art performance in event relation reasoning. \textbf{This suggests that an increased number of frames necessitates more accurate mechanisms for event relation understanding, whereas models with limited frame input tend to struggle with event relation reasoning.}

\textbf{Bias Analysis: }In Figure \ref{bias_analysis}, we evaluate several models under the QA settings to assess their reliance on language bias and vision\&language bias. As the results show, \textbf{vision\&language bias is more pronounced than language bias.} VideoLLMs often fail to capture key visual cues when performing event relation reasoning, leading to relation-level hallucinations. This observation is consistent with the conclusions drawn from the visualization result in Figure \ref{visuallll}. Subsequently, while VideoLLaMA3-7B exhibits a relatively lower degree of hallucination, its performance on VERHallu-QA and VERHallu-CFQA remains poor. This suggests that VideoLLaMA3-7B frequently opts for ambiguous responses such as `cannot understand', reflecting its limited capacity for event relation comprehension. In contrast, LLaVA-NeXT-7B-DPO shows the most severe hallucination issues. Notably, QwenVL-2.5-7B demonstrates the least susceptibility to hallucinations and achieves the best overall performance on VERHallu.

\begin{table}
\centering
\caption{Performance of KFP across different layers (\%).}
\label{tab:layer_performance_percent}
\begin{tabular}{l cccc}
\toprule
\multirow{2}{*}{\textbf{Layer}} & 
\multicolumn{4}{c}{\textbf{VERHallu}} \\
\cmidrule(lr){2-5}
& \textbf{CFQA} & \textbf{QA-Causal} & \textbf{QA-Temporal} & \textbf{QA-Subevent} \\
\midrule
0--5   & 50.1 & 39.2 & 49.1 & 41.5 \\
0--10  & 49.0 & 41.0 & 48.6 & 41.5 \\
5--10  & 44.1 & 41.4 & 49.5 & 41.1 \\
5--15  & 43.5 & 42.1 & 49.1 & 45.3 \\
10--15 & 45.4 & 43.3 & 51.9 & 45.0 \\
10--20 & 47.5 & 41.2 & 49.1 & 42.2 \\
15--20 & 50.2 & 39.2 & 44.8 & 39.5 \\
15--25 & 51.4 & 38.8 & 45.8 & 38.4 \\
20--25 & 50.6 & 39.4 & 49.5 & 41.1 \\
\midrule
Baseline & 50.6 & 40.8 & 50.5 & 40.7 \\
\bottomrule
\end{tabular}
\end{table}

\subsection{Results with KFP}
In the Table \ref{tab:models_kfp}, we incorporate the proposed KFP method into LLaVA-NeXT-7B and QwenVL-2.5-7B for comparative evaluation. Additionally, we include the representative Visual Contrastive Decoding (VCD) approach for baseline comparison. Due to the model's limited capabilities, introducing the VCD method in LLaVA-NeXT-7B significantly degraded its instruction-following ability, rendering it unable to complete the tasks. Therefore, we did not report its results in the table.

\subsubsection{Main Results}
From the comparison, we observe that the Visual Contrastive Decoding (VCD) strategy \cite{vcd} is not well-suited for addressing event relation hallucination. In contrast, the KFP algorithm consistently improves the performance of LLaVA-NeXT-7B and significantly improves certain aspects of QwenVL-2.5-7B, such as QA-Subevent. In LLaVA-NeXT-7B, which has relatively weaker reasoning abilities, KFP method achieves an overall and significant improvement in event relation understanding (e.g., RC-S +39.1\%). However, when applied to the stronger reasoning model QwenVL-2.5-7B, its performance on CFQA is negatively affected (CFQA -6.0\%). As shown in the Table \ref{tab:layer_performance_percent}, excluding CFQA, KFP is most effective in the middle layers (10–15), where multimodal fusion occurs, while shallow or deep layers yield minimal impact, consistent with \cite{yin2025clearsight}. In CFQA, since there is no critical visual information requiring attention, attention enhancement in early layers can be corrected during subsequent fusion, and late-layer enhancement introduces negligible interference once multimodal integration has been completed. In contrast, applying KFP to the middle layers amplifies semantically insignificant visual features in counterfactual scenarios, ultimately degrading performance. This is a major limitation of KFP. Moreover, KFP shows a favorable inference speed performance and can significantly enhance the model's ability to understand the overall video event relations (SRH).

\begin{figure*}
  \centering
\includegraphics[width=0.65\textwidth]{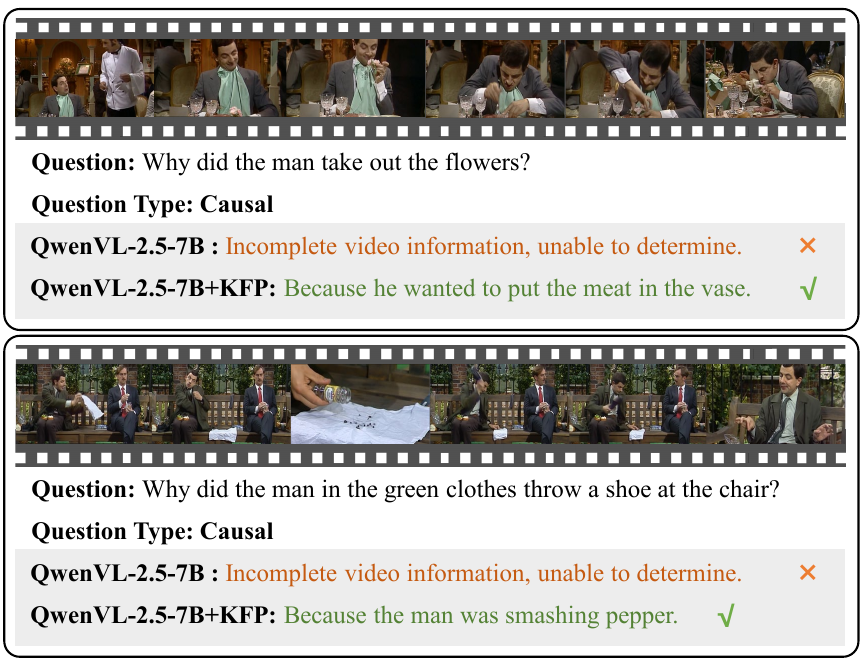} 
  \caption{Case study of our proposed KFP.} 
  \label{case}
\end{figure*}

\subsubsection{Parameter Study}

\begin{table}
\centering
 \caption{Ablation Study of $m$ in VERHallu with LLaVA-NeXT-7B.}
\begin{tabular}{l cccc}
\toprule
\multirow{2}{*}{\textbf{m}}   & 
\multicolumn{3}{c}{\textbf{VERHallu-QA}} \\
\cmidrule(lr){2-4}
 &  \textbf{Causal} & \textbf{Temporal} & \textbf{Subevent} \\
\midrule
Base &15.0 &15.0 &12.0 \\
2 &14.9 &14.2 &12.7 \\
3 &14.5 &14.1 &11.6 \\
4 &14.5 &15.1 &12.4 \\
5 &15.3 &19.3 &15.1 \\
6 &14.7 &15.6 &12.8 \\
\bottomrule
\end{tabular}
\label{tab:models_kfp_m}
\end{table}

\begin{table}
\centering
\caption{Ablation Study of $\beta$ in VERHallu with LLaVA-NeXT-7B.}
\begin{tabular}{l cccc}
\toprule
\multirow{2}{*}{\textbf{$\beta$}}   & 
\multicolumn{3}{c}{\textbf{VERHallu-QA}} \\
\cmidrule(lr){2-4}
 &  \textbf{Causal} & \textbf{Temporal} & \textbf{Subevent} \\
\midrule
Base &15.0 &15.0 &12.0 \\
0.55 &15.1 &16.0 &14.3 \\
0.60 &15.3 &19.3 &15.1 \\
0.65 &15.1 &15.2 &14.3 \\
0.70 &15.3 &15.6 &14.3 \\
0.75 &15.1 &15.1 &12.7 \\
\bottomrule
\end{tabular}
\label{tab:models_kfp_beta}
\end{table}

In the KFP algorithm, we conducted ablation studies on the hyperparameters $m$ and $\beta$, as shown in the Table \ref{tab:models_kfp_m} and  \ref{tab:models_kfp_beta}. From the table, it can be observed that the best performance is achieved when $m$ is set to 5 and $\beta$ is set to 0.6.

The number of layers and the parameter $\sigma$ in Gaussian weighting were kept fixed, following previous work \cite{yin2025clearsight}, with the number of layers set between 8 and 15 and $\sigma$ set to 1.

\subsubsection{Case Study}

To provide a more intuitive illustration of the characteristics of the KFP method, we conduct case studies. As shown in Figure \ref{case}, the KFP method captures a greater number of events in both examples, establishing precise causal relations such as between `taking out the flowers’ and `putting the meat into the vase’, as well as between `crushing the pepper’ and `hitting the chair with a shoe’. In contrast, the non-KFP method tends to be overconfident, neglecting surrounding events and consequently producing incorrect results. Overall, KFP effectively enhances the model’s ability to capture event relations, leading to more comprehensive and accurate answers.

\subsubsection{Limitation Analysis}
However, the issue of event relation hallucination is not merely a deficiency in capturing individual events; rather, it reflects a deeper challenge in comprehensively understanding the relations between events based on their evolving details. This necessitates not only robust visual perception capabilities but also precise and holistic reasoning abilities within the model. Although our proposed KFP algorithm enhances the model's attention to visual features to some extent, it still lacks the visual perception capacity explicitly oriented toward relation reasoning, as well as a strong ability to comprehend complex event relations. Therefore, our method should be regarded as an exploratory step toward addressing this challenge. Future efforts should focus on enhancing VideoLLMs via improved architectures and training data tailored for event relation reasoning.

\section{Conclusion}

In this paper, we introduce VERHallu, a novel benchmark designed to evaluate the event relation hallucination through relation classification, question answering, and counterfactual question answering tasks. Our analysis shows that state-of-the-art VideoLLMs often rely on prior knowledge and partial frame cues, leading to inaccurate reasoning across events. To mitigate this, we propose the Key-Frame Propagating (KFP) strategy, which enhances multi-event understanding by reallocating frame-level attention within intermediate layers. Experiments demonstrate that KFP effectively reduces the event relation hallucination without compromising inference speed. We hope our work will encourage further research into tackling event-level reasoning challenges and inspire more effective approaches to mitigating the event relation hallucination in VideoLLMs.

\section{ACKNOWLEDGEMENTS}

This work is supported by the National Natural Science Foundation of China [62576149] and the Fundamental Research Funds for the Central University, JLU.

\bibliographystyle{IEEEtran}
\bibliography{eg.bib}

@article{ji2023survey,
  title={Survey of hallucination in natural language generation},
  author={Ji, Ziwei and Lee, Nayeon and Frieske, Rita and Yu, Tiezheng and Su, Dan and Xu, Yan and Ishii, Etsuko and Bang, Ye Jin and Madotto, Andrea and Fung, Pascale},
  journal={ACM computing surveys},
  volume={55},
  number={12},
  pages={1--38},
  year={2023},
  publisher={ACM New York, NY}
}

@inproceedings{li2025vidhalluc,
  title={Vidhalluc: Evaluating temporal hallucinations in multimodal large language models for video understanding},
  author={Li, Chaoyu and Im, Eun Woo and Fazli, Pooyan},
  booktitle={Proceedings of the Computer Vision and Pattern Recognition Conference},
  pages={13723--13733},
  year={2025}
}

@article{wang2024videohallucer,
  title={Videohallucer: Evaluating intrinsic and extrinsic hallucinations in large video-language models},
  author={Wang, Yuxuan and Wang, Yueqian and Zhao, Dongyan and Xie, Cihang and Zheng, Zilong},
  journal={arXiv preprint arXiv:2406.16338},
  year={2024}
}

@article{li2025videohallu,
  title={VideoHallu: Evaluating and Mitigating Multi-modal Hallucinations on Synthetic Video Understanding},
  author={Li, Zongxia and Wu, Xiyang and Shi, Guangyao and Qin, Yubin and Du, Hongyang and Zhou, Tianyi and Manocha, Dinesh and Boyd-Graber, Jordan Lee},
  journal={arXiv preprint arXiv:2505.01481},
  year={2025}
}

@article{oquab2023dinov2,
  title={Dinov2: Learning robust visual features without supervision},
  author={Oquab, Maxime and Darcet, Timoth{\'e}e and Moutakanni, Th{\'e}o and Vo, Huy and Szafraniec, Marc and Khalidov, Vasil and Fernandez, Pierre and Haziza, Daniel and Massa, Francisco and El-Nouby, Alaaeldin and others},
  journal={arXiv preprint arXiv:2304.07193},
  year={2023}
}

@inproceedings{bae2025mash,
  title={MASH-VLM: Mitigating Action-Scene Hallucination in Video-LLMs through Disentangled Spatial-Temporal Representations},
  author={Bae, Kyungho and Kim, Jinhyung and Lee, Sihaeng and Lee, Soonyoung and Lee, Gunhee and Choi, Jinwoo},
  booktitle={Proceedings of the Computer Vision and Pattern Recognition Conference},
  pages={13744--13753},
  year={2025}
}

@inproceedings{kong2025mhbench,
  title={MHBench: Demystifying Motion Hallucination in VideoLLMs},
  author={Kong, Ming and Zeng, Xianzhou and Chen, Luyuan and Li, Yadong and Yan, Bo and Zhu, Qiang},
  booktitle={Proceedings of the AAAI Conference on Artificial Intelligence},
  volume={39},
  number={4},
  pages={4401--4409},
  year={2025}
}

@inproceedings{vcd,
  title={Mitigating object hallucinations in large vision-language models through visual contrastive decoding},
  author={Leng, Sicong and Zhang, Hang and Chen, Guanzheng and Li, Xin and Lu, Shijian and Miao, Chunyan and Bing, Lidong},
  booktitle={Proceedings of the IEEE/CVF Conference on Computer Vision and Pattern Recognition},
  pages={13872--13882},
  year={2024}
}

@article{yu2024prompting,
  title={Prompting video-language foundation models with domain-specific fine-grained heuristics for video question answering},
  author={Yu, Ting and Fu, Kunhao and Wang, Shuhui and Huang, Qingming and Yu, Jun},
  journal={IEEE Transactions on Circuits and Systems for Video Technology},
  year={2024},
  publisher={IEEE}
}

@article{wang2025align,
  title={Align is not enough: Multimodal universal jailbreak attack against multimodal large language models},
  author={Wang, Youze and Hu, Wenbo and Dong, Yinpeng and Liu, Jing and Zhang, Hanwang and Hong, Richang},
  journal={IEEE Transactions on Circuits and Systems for Video Technology},
  year={2025},
  publisher={IEEE}
}

@article{fang2024linguistic,
  title={Linguistic hallucination for text-based video retrieval},
  author={Fang, Sheng and Dang, Tiantian and Wang, Shuhui and Huang, Qingming},
  journal={IEEE Transactions on Circuits and Systems for Video Technology},
  volume={34},
  number={10},
  pages={9692--9705},
  year={2024},
  publisher={IEEE}
}

@article{bubeck2023sparks,
  title={Sparks of artificial general intelligence: Early experiments with gpt-4},
  author={Bubeck, S{\'e}bastien and Chandrasekaran, Varun and Eldan, Ronen and Gehrke, Johannes and Horvitz, Eric and Kamar, Ece and Lee, Peter and Lee, Yin Tat and Li, Yuanzhi and Lundberg, Scott and others},
  journal={arXiv preprint arXiv:2303.12712},
  year={2023}
}

@article{team2023gemini,
  title={Gemini: a family of highly capable multimodal models},
  author={Team, Gemini and Anil, Rohan and Borgeaud, Sebastian and Alayrac, Jean-Baptiste and Yu, Jiahui and Soricut, Radu and Schalkwyk, Johan and Dai, Andrew M and Hauth, Anja and Millican, Katie and others},
  journal={arXiv preprint arXiv:2312.11805},
  year={2023}
}

@inproceedings{yin2025clearsight,
  title={ClearSight: Visual Signal Enhancement for Object Hallucination Mitigation in Multimodal Large Language Models},
  author={Yin, Hao and Si, Guangzong and Wang, Zilei},
  booktitle={Proceedings of the Computer Vision and Pattern Recognition Conference},
  pages={14625--14634},
  year={2025}
}

@inproceedings{li2024mvbench,
  title={Mvbench: A comprehensive multi-modal video understanding benchmark},
  author={Li, Kunchang and Wang, Yali and He, Yinan and Li, Yizhuo and Wang, Yi and Liu, Yi and Wang, Zun and Xu, Jilan and Chen, Guo and Luo, Ping and others},
  booktitle={Proceedings of the IEEE/CVF Conference on Computer Vision and Pattern Recognition},
  pages={22195--22206},
  year={2024}
}

@inproceedings{lin2024video,
  title={Video-LLaVA: Learning United Visual Representation by Alignment Before Projection},
  author={Lin, Bin and Ye, Yang and Zhu, Bin and Cui, Jiaxi and Ning, Munan and Jin, Peng and Yuan, Li},
  booktitle={Proceedings of the 2024 Conference on Empirical Methods in Natural Language Processing},
  pages={5971--5984},
  year={2024}
}

@article{wang2022maven,
  title={Maven-ere: A unified large-scale dataset for event coreference, temporal, causal, and subevent relation extraction},
  author={Wang, Xiaozhi and Chen, Yulin and Ding, Ning and Peng, Hao and Wang, Zimu and Lin, Yankai and Han, Xu and Hou, Lei and Li, Juanzi and Liu, Zhiyuan and others},
  journal={arXiv preprint arXiv:2211.07342},
  year={2022}
}

@article{sun2024hallucination,
  title={Hallucination mitigation prompts long-term video understanding},
  author={Sun, Yiwei and Liu, Zhihang and Liu, Chuanbin and Pu, Bowei and Zhang, Zhihan and Xie, Hongtao},
  journal={arXiv preprint arXiv:2406.11333},
  year={2024}
}

@article{Eventhallusion,
  title={Eventhallusion: Diagnosing event hallucinations in video llms},
  author={Zhang, Jiacheng and Jiao, Yang and Chen, Shaoxiang and Zhao, Na and Tan, Zhiyu and Li, Hao and Chen, Jingjing},
  journal={arXiv preprint arXiv:2409.16597},
  year={2024}
}

@article{sahoo2024comprehensive,
  title={A comprehensive survey of hallucination in large language, image, video and audio foundation models},
  author={Sahoo, Pranab and Meharia, Prabhash and Ghosh, Akash and Saha, Sriparna and Jain, Vinija and Chadha, Aman},
  journal={arXiv preprint arXiv:2405.09589},
  year={2024}
}

@article{huang2024visual,
  title={Visual hallucinations of multi-modal large language models},
  author={Huang, Wen and Liu, Hongbin and Guo, Minxin and Gong, Neil Zhenqiang},
  journal={arXiv preprint arXiv:2402.14683},
  year={2024}
}

@article{zhou2022learning,
  title={Learning to prompt for vision-language models},
  author={Zhou, Kaiyang and Yang, Jingkang and Loy, Chen Change and Liu, Ziwei},
  journal={International Journal of Computer Vision},
  volume={130},
  number={9},
  pages={2337--2348},
  year={2022},
  publisher={Springer}
}

@inproceedings{guan2024hallusionbench,
  title={Hallusionbench: an advanced diagnostic suite for entangled language hallucination and visual illusion in large vision-language models},
  author={Guan, Tianrui and Liu, Fuxiao and Wu, Xiyang and Xian, Ruiqi and Li, Zongxia and Liu, Xiaoyu and Wang, Xijun and Chen, Lichang and Huang, Furong and Yacoob, Yaser and others},
  booktitle={Proceedings of the IEEE/CVF Conference on Computer Vision and Pattern Recognition},
  pages={14375--14385},
  year={2024}
}

@inproceedings{li2023evaluating,
  title={Evaluating Object Hallucination in Large Vision-Language Models},
  author={Li, Yifan and Du, Yifan and Zhou, Kun and Wang, Jinpeng and Zhao, Wayne Xin and Wen, Ji-Rong},
  booktitle={Proceedings of the 2023 Conference on Empirical Methods in Natural Language Processing},
  pages={292--305},
  year={2023}
}

@article{touvron2023llama,
  title={Llama 2: Open foundation and fine-tuned chat models},
  author={Touvron, Hugo and Martin, Louis and Stone, Kevin and Albert, Peter and Almahairi, Amjad and Babaei, Yasmine and Bashlykov, Nikolay and Batra, Soumya and Bhargava, Prajjwal and Bhosale, Shruti and others},
  journal={arXiv preprint arXiv:2307.09288},
  year={2023}
}

@article{zhu2023minigpt,
  title={Minigpt-4: Enhancing vision-language understanding with advanced large language models},
  author={Zhu, Deyao and Chen, Jun and Shen, Xiaoqian and Li, Xiang and Elhoseiny, Mohamed},
  journal={arXiv preprint arXiv:2304.10592},
  year={2023}
}

@inproceedings{ma2024vista,
  title={Vista-llama: Reducing hallucination in video language models via equal distance to visual tokens},
  author={Ma, Fan and Jin, Xiaojie and Wang, Heng and Xian, Yuchen and Feng, Jiashi and Yang, Yi},
  booktitle={Proceedings of the IEEE/CVF Conference on Computer Vision and Pattern Recognition},
  pages={13151--13160},
  year={2024}
}

@article{ding2025pami,
  title={PaMi-VDPO: Mitigating Video Hallucinations by Prompt-Aware Multi-Instance Video Preference Learning},
  author={Ding, Xinpeng and Zhang, Kui and Han, Jianhua and Hong, Lanqing and Xu, Hang and Li, Xiaomeng},
  journal={arXiv preprint arXiv:2504.05810},
  year={2025}
}

@article{gao2025exploring,
  title={Exploring hallucination of large multimodal models in video understanding: Benchmark, analysis and mitigation},
  author={Gao, Hongcheng and Qu, Jiashu and Tang, Jingyi and Bi, Baolong and Liu, Yue and Chen, Hongyu and Liang, Li and Su, Li and Huang, Qingming},
  journal={arXiv preprint arXiv:2503.19622},
  year={2025}
}

@article{li2025token,
  title={Token Activation Map to Visually Explain Multimodal LLMs},
  author={Li, Yi and Wang, Hualiang and Ding, Xinpeng and Wang, Haonan and Li, Xiaomeng},
  journal={arXiv preprint arXiv:2506.23270},
  year={2025}
}

@article{chen2024sharegpt4video,
  title={Sharegpt4video: Improving video understanding and generation with better captions},
  author={Chen, Lin and Wei, Xilin and Li, Jinsong and Dong, Xiaoyi and Zhang, Pan and Zang, Yuhang and Chen, Zehui and Duan, Haodong and Tang, Zhenyu and Yuan, Li and others},
  journal={Advances in Neural Information Processing Systems},
  volume={37},
  pages={19472--19495},
  year={2024}
}

@article{yao2024minicpm,
  title={Minicpm-v: A gpt-4v level mllm on your phone},
  author={Yao, Yuan and Yu, Tianyu and Zhang, Ao and Wang, Chongyi and Cui, Junbo and Zhu, Hongji and Cai, Tianchi and Li, Haoyu and Zhao, Weilin and He, Zhihui and others},
  journal={arXiv preprint arXiv:2408.01800},
  year={2024}
}

@article{ye2024mplug,
  title={mplug-owl3: Towards long image-sequence understanding in multi-modal large language models},
  author={Ye, Jiabo and Xu, Haiyang and Liu, Haowei and Hu, Anwen and Yan, Ming and Qian, Qi and Zhang, Ji and Huang, Fei and Zhou, Jingren},
  journal={arXiv preprint arXiv:2408.04840},
  year={2024}
}

@article{li2024llava,
  title={Llava-next-interleave: Tackling multi-image, video, and 3d in large multimodal models},
  author={Li, Feng and Zhang, Renrui and Zhang, Hao and Zhang, Yuanhan and Li, Bo and Li, Wei and Ma, Zejun and Li, Chunyuan},
  journal={arXiv preprint arXiv:2407.07895},
  year={2024}
}

@article{Llava-onevision,
  title={Llava-onevision: Easy visual task transfer},
  author={Li, Bo and Zhang, Yuanhan and Guo, Dong and Zhang, Renrui and Li, Feng and Zhang, Hao and Zhang, Kaichen and Zhang, Peiyuan and Li, Yanwei and Liu, Ziwei and others},
  journal={arXiv preprint arXiv:2408.03326},
  year={2024}
}

@article{zhang2025videollama,
  title={Videollama 3: Frontier multimodal foundation models for image and video understanding},
  author={Zhang, Boqiang and Li, Kehan and Cheng, Zesen and Hu, Zhiqiang and Yuan, Yuqian and Chen, Guanzheng and Leng, Sicong and Jiang, Yuming and Zhang, Hang and Li, Xin and others},
  journal={arXiv preprint arXiv:2501.13106},
  year={2025}
}

@article{li2025videochat,
  title={Videochat-r1: Enhancing spatio-temporal perception via reinforcement fine-tuning},
  author={Li, Xinhao and Yan, Ziang and Meng, Desen and Dong, Lu and Zeng, Xiangyu and He, Yinan and Wang, Yali and Qiao, Yu and Wang, Yi and Wang, Limin},
  journal={arXiv preprint arXiv:2504.06958},
  year={2025}
}

@article{liu2024survey,
  title={A survey on hallucination in large vision-language models},
  author={Liu, Hanchao and Xue, Wenyuan and Chen, Yifei and Chen, Dapeng and Zhao, Xiutian and Wang, Ke and Hou, Liping and Li, Rongjun and Peng, Wei},
  journal={arXiv preprint arXiv:2402.00253},
  year={2024}
}

@inproceedings{chen2024internvl,
  title={Internvl: Scaling up vision foundation models and aligning for generic visual-linguistic tasks},
  author={Chen, Zhe and Wu, Jiannan and Wang, Wenhai and Su, Weijie and Chen, Guo and Xing, Sen and Zhong, Muyan and Zhang, Qinglong and Zhu, Xizhou and Lu, Lewei and others},
  booktitle={Proceedings of the IEEE/CVF conference on computer vision and pattern recognition},
  pages={24185--24198},
  year={2024}
}

@article{wang2024qwen2,
  title={Qwen2-vl: Enhancing vision-language model's perception of the world at any resolution},
  author={Wang, Peng and Bai, Shuai and Tan, Sinan and Wang, Shijie and Fan, Zhihao and Bai, Jinze and Chen, Keqin and Liu, Xuejing and Wang, Jialin and Ge, Wenbin and others},
  journal={arXiv preprint arXiv:2409.12191},
  year={2024}
}

@inproceedings{li2023blip,
  title={Blip-2: Bootstrapping language-image pre-training with frozen image encoders and large language models},
  author={Li, Junnan and Li, Dongxu and Savarese, Silvio and Hoi, Steven},
  booktitle={International conference on machine learning},
  pages={19730--19742},
  year={2023},
  organization={PMLR}
}

@article{bai2025qwen2,
  title={Qwen2. 5-vl technical report},
  author={Bai, Shuai and Chen, Keqin and Liu, Xuejing and Wang, Jialin and Ge, Wenbin and Song, Sibo and Dang, Kai and Wang, Peng and Wang, Shijie and Tang, Jun and others},
  journal={arXiv preprint arXiv:2502.13923},
  year={2025}
}

@article{gpt,
  title={Gpt-4 technical report},
  author={Achiam, Josh and Adler, Steven and Agarwal, Sandhini and Ahmad, Lama and Akkaya, Ilge and Aleman, Florencia Leoni and Almeida, Diogo and Altenschmidt, Janko and Altman, Sam and Anadkat, Shyamal and others},
  journal={arXiv preprint arXiv:2303.08774},
  year={2023}
}

@article{liu2023visual,
  title={Visual instruction tuning},
  author={Liu, Haotian and Li, Chunyuan and Wu, Qingyang and Lee, Yong Jae},
  journal={Advances in neural information processing systems},
  volume={36},
  pages={34892--34916},
  year={2023}
}

@article{llama,
  title={The llama 3 herd of models},
  author={Grattafiori, Aaron and Dubey, Abhimanyu and Jauhri, Abhinav and Pandey, Abhinav and Kadian, Abhishek and Al-Dahle, Ahmad and Letman, Aiesha and Mathur, Akhil and Schelten, Alan and Vaughan, Alex and others},
  journal={arXiv preprint arXiv:2407.21783},
  year={2024}
}

@inproceedings{qin2025question,
  title={Question-answering dense video events},
  author={Qin, Hangyu and Xiao, Junbin and Yao, Angela},
  booktitle={Proceedings of the 48th International ACM SIGIR Conference on Research and Development in Information Retrieval},
  pages={884--894},
  year={2025}
}

@inproceedings{lin2024towards,
  title={Towards explainable harmful meme detection through multimodal debate between large language models},
  author={Lin, Hongzhan and Luo, Ziyang and Gao, Wei and Ma, Jing and Wang, Bo and Yang, Ruichao},
  booktitle={Proceedings of the ACM Web Conference 2024},
  pages={2359--2370},
  year={2024}
}

@article{qwen,
  title={Qwen3 technical report},
  author={Yang, An and Li, Anfeng and Yang, Baosong and Zhang, Beichen and Hui, Binyuan and Zheng, Bo and Yu, Bowen and Gao, Chang and Huang, Chengen and Lv, Chenxu and others},
  journal={arXiv preprint arXiv:2505.09388},
  year={2025}
}

@inproceedings{luo2025imagescope,
  title={ImageScope: Unifying Language-Guided Image Retrieval via Large Multimodal Model Collective Reasoning},
  author={Luo, Pengfei and Zhou, Jingbo and Xu, Tong and Xia, Yuan and Xu, Linli and Chen, Enhong},
  booktitle={Proceedings of the ACM on Web Conference 2025},
  pages={1666--1682},
  year={2025}
}












\newpage

 




\vfill

\end{document}